\pdfoutput=1

\documentclass[11pt]{article} 

\usepackage{emnlp2021}

\usepackage{times}
\usepackage{latexsym}

\usepackage[T1]{fontenc}

\usepackage[inline]{enumitem}
\usepackage{microtype}
\usepackage{amsfonts}
\usepackage{bm}
\usepackage{array,amsmath}
\usepackage{amssymb}
\usepackage{dsfont}
\usepackage{float}
\usepackage{graphicx}
\usepackage{algorithm}
\usepackage{algorithmicx}
\usepackage{algpseudocode}
\usepackage{pgf,tikz}
\usepackage{xcolor}
\usepackage{color, colortbl}
\usepackage{bm}
\usepackage{mathrsfs}
\usepackage{multirow}
\usepackage{booktabs}
\usetikzlibrary{arrows}
\usepackage{threeparttable}
\usepackage[inline]{enumitem}
\usepackage{endnotes}
\usepackage[normalem]{ulem}

\let\times\relax
\DeclareMathSymbol{\times}{\mathbin}{symbols}{"02}

\def\tabref#1{Table~\ref{tab:#1}} \def\tablabel#1{\label{tab:#1}\label{p:#1}}
 
 \def\eqref#1{Eq.~\ref{eqn:#1}}

\newcounter{notecounter} 
\newcommand{\enotesoff}{\long\gdef\enote##1##2{}}
\newcommand{\enoteson}{\long\gdef\enote##1##2{{ \stepcounter{notecounter}
			{\large\bf \hspace{1cm}\arabic{notecounter} $<<<$ ##1: ##2 $>>>$\hspace{1cm}}}}}
\enoteson \enotesoff \long \def\eat#1{}

 \def\simalign{SimAlign} 
\def\nlangs{1334} \def\neditions{1758} 
\def\nversesexact{20,470,892} \def\avgverses{11,520} 
 \def\avgtokens{28.6} \def\eflomal{Eflomal}
\def\framework{MPWA} 

\enoteson
\enotesoff

\usepackage[utf8]{inputenc}

\usepackage{microtype}

\title{Graph Algorithms for Multiparallel Word Alignment}

\author{Ayyoob Imani\thanks{\mbox{\ \ } Equal contribution.}$\ \,^1$,
	Masoud Jalili Sabet$^{*}$$^1$, L\"{u}tfi Kerem \c{S}enel$^1$, \\
	{\bf Philipp Dufter$^1$, Fran\c{c}ois Yvon$^2$, Hinrich Sch\"{u}tze$^1$}\\
	$^1$Center for Information and Language Processing (CIS), LMU Munich, Germany\\
	$^2$Universit\'{e} Paris-Saclay, CNRS, LISN, France\\
	{\tt \{ayyoob, masoud, lksenel, philipp\}@cis.lmu.de,} \\
	{\tt francois.yvon@limsi.fr}
}

\begin{document}

\maketitle 
\begin{abstract}
With the advent of end-to-end deep learning approaches in machine translation,
interest in word alignments initially decreased; however, they have again
become a focus of research more recently.
Alignments are useful for typological research, transferring formatting
like markup to translated texts, and can be used in the decoding of machine
translation systems. At the same time, massively multilingual processing is 
becoming an important NLP scenario, and pretrained language and machine
translation models that are truly multilingual are proposed. However, 
most alignment algorithms rely on bitexts only and do not leverage the fact
that many parallel corpora are multiparallel. In this work, we exploit the 
multiparallelity of corpora by representing an initial set of bilingual 
alignments as a graph and then predicting additional edges in the graph.
We present two graph algorithms for edge prediction: one inspired by 
recommender systems and one based on network link prediction. 
Our experimental results show absolute improvements in $F_1$
of up to
28\% 
over the baseline bilingual word aligner in
different datasets.
\end{abstract}

\section{Introduction}

\begin{figure}[h!]
	\centering
	\includegraphics[width=1.0\linewidth]{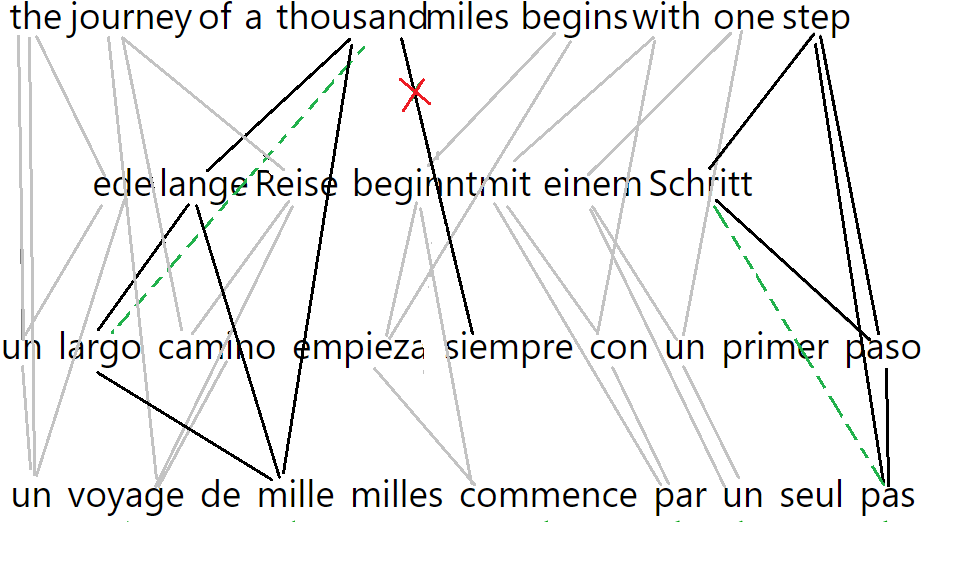}
	\caption{ 
Bilingual alignments of a verse in English, German, Spanish, and French. 
Two of the alignment edges not found by the bilingual method are
German ``Schritt'' to French ``pas''
and Spanish ``largo'' to English ``thousand miles''.
By looking at the structure of the entire graph, one
	can infer the correctness of these two edges.}
\label{fig:lexdiff} 
\end{figure}

Word alignment is a challenging NLP task that
plays an essential role in statistical machine translation 
and is useful for neural machine translation
 \cite{alkhouli-ney-2017-biasing, alkhouli-etal-2016-alignment, koehn-etal-2003-statistical}.
 Other applications of word alignments include bilingual lexicon induction, annotation projection, 
 and typological analysis  \cite{shi-etal-2021-bilingual, rasooli2018cross, muller-2017-treatment, 
 lewis-xia-2008-automatically}.
With the advent of deep learning, interest in word alignment
 initially decreased.
However, recently a new wave of publications has again drawn attention
to the task
 \cite{jalili-sabet-etal-2020-simalign, dou-neubig-2021-word, marchisio2021embedding, wu-dredze-2020-explicit}.

In this paper we propose \emph{\framework{}} (MultiParallel Word
Alignment), a framework that
employs graph algorithms to
  exploit the information latent in a multiparallel corpus to achieve better
  word alignments than aligning pairs of languages in isolation.
  Starting from translations of a sentence in multiple languages in a multiparallel corpus,
  \framework{} generates bilingual word alignments for all
  language pairs
using
  any available bilingual word aligner. \framework{} then
  improves the quality of word alignments for 
  a target language pair by inspecting how they are
 aligned to other languages.
\emph{The central idea is to exploit the graph structure of an
 initial multiparallel word alignment
 to improve the alignment for a target language pair.}
To
 this end, \framework{} casts the multiparallel word alignment task as a 
 link (or edge) prediction problem. We explore standard
algorithms for this purpose:
Adamic-Adar and matrix factorization. While these two graph-based
 algorithms are quite  different  and are used in different applications,  we will 
  show that \framework{} effectively leverages them for
 high-performing word alignment.

Link prediction methods are used to predict whether there should be a link between two nodes in a 
graph. They have various applications like movie recommendations, knowledge graph completion, and
metabolic network reconstruction  \cite{zhang2018link}. We
use the Adamic-Adar index  \cite{adamic2003friends}; it is a second-order 
link prediction algorithm, i.e., it exploits the information
of neighbors
that are up to two hops aways from the starting target nodes
 \cite{zhou2009predicting}. We use a second-order algorithm since a set of aligned words 
in multiple languages 
 (representing a concept) tends to establish a clique  \cite{dufter-etal-2018-embedding}. This means
  that exploring the influence of nodes at a distance of two
  in the graph 
 provides informative signals while at the same time keeping runtime complexity low.

Matrix factorization is a collaborative filtering algorithm
that is most prominently  used in recommender systems
where it provides users with product recommendations based on their interactions
with other products. This method is especially useful if the
matrix is
sparse \cite{koren2009matrix}. This is true for our application: Given two
translations of a sentence with lengths \(M\) and \(N\), among all possible alignment links
 (\(M \times N\)), only a few (\(O(M+N)\)) are correct. This
is partly due to fertility: words in the
 source language generally have only a few possible matches in the
 target language
  \cite{zhao-gildea-2010-fast}.

 A multiparallel corpus provides parallel sentences in more than two languages. This type of 
 corpus facilitates the study of multiple languages together,
 which is
especially 
important 
 for research on low resource languages. As far as we know, out of all available multiparallel corpora, 
 the Parallel Bible Corpus  \cite{mayer-cysouw-2014-creating} (PBC) provides the highest language coverage, supporting
 \nlangs{} different languages, many of which belong to categories 
 0 and 1  \cite{joshi-etal-2020-state} -- that is, they are
 languages
 for which no language 
 technologies are available and that are severely
 underresourced.

\framework{} has especially strong word alignment improvements for distant
language pairs for which existing bilingual word aligners perform poorly.
Much work that addresses low resource languages relies on
the availabiliy of monolingual corpora.
Complementarily, \framework{}
 assumes the existence of a very small (a few 10,000s of sentences
 in our case) parallel corpus and
then takes advantage of
information from 
 the other languages in the parallel corpus. This is
an alternative approach that is
especially important for  low
 resource languages for which monolingual data often are not available.
 
The PBC corpus does not contain a word alignment gold
standard. To conduct the comparative evaluation of our new method,
we port three existing word alignment 
 gold standards of Bible 
 translations to PBC, for the language pairs 
English-French, Finnish-Hebrew and Finnish-Greek. We also create artificial multiparallel
 datasets for four widely used word alignment datasets using machine translation.
 We evaluate our method with all seven datasets. Results demonstrate substantial
 improvements in all scenarios.

Our main contributions are: 
\begin{enumerate}
	\item We propose
two graph-based algorithms for link prediction (i.e., the
	prediction of word alignment edges in the alignment graph),
one based on second-order link prediction
and one based on recommender systems 
	 for improving word alignment in a multiparallel corpus and show that they perform better than established baselines. 
	\item We port and publish three word alignment gold
	standards for the Parallel Bible Corpus.
	\item We show that our method is also applicable, using machine translation,
to scenarios where multiparallel data is not available.
	\item We publish our code\footnote{\url{https://github.com/cisnlp/graph-align}} and data.
\end{enumerate}

\section{Related Work}

\textbf{Bilingual Word Aligners} 
take different	approaches. Some are based
	on statistical analysis, like IBM models
	 \cite{brown1993mathematics}, Giza++  \cite{och03:asc}, fast-align
	 \cite{dyer-etal-2013-simple} and Eflomal  \cite{ostling2016efficient}.
	Another more recent group, 
	including SimAlign
	 \cite{jalili-sabet-etal-2020-simalign} and Awesome-align  \cite{dou-neubig-2021-word},
utilizes neural language models.	The last group is based on neural machine translation  \cite{garg-etal-2019-jointly,zenkel-etal-2020-end}.
	While neural models outperform statistical models, 
	for cases where only a small parallel dataset is available, statistical models are still
	superior. In this paper we use PBC, a corpus
	with \nlangs{} languages,
of which only about
	two hundred are supported by multilingual language models like Bert and XLM-R 
	 \cite{devlin-etal-2019-bert, conneau-etal-2020-unsupervised}.
	\framework{} can
leverage multiparallelism 
on top of any bilingual word aligner; in this paper, 
	we use Eflomal and \simalign{}.
	
\textbf{Multiparallel corpus alignment.}
	Most work on word alignment has focused on bilingual
	corpora. To the best of our knowledge,
	only one method
	specifically designed for multiparallel corpora
	was previously proposed:
	 \cite{ostling2014bayesian}.\footnote{\url{https://github.com/robertostling/eflomal}}
	However this method is outperformed by a
	``biparallel'' method by the same author, 
	\eflomal{}  \cite{ostling2016efficient}. We
	compare with \eflomal{} in our experiments.
	
	 \citet{cohn-lapata-2007-machine} make use of multiparallel corpora to obtain more reliable
	translations from small datasets.
	 \citet{kumar-etal-2007-improving} show that multiparallel corpora
	can be of benefit  to reach 
	better performance in phrase-based statistical machine translation (SMT).
	 \citet{filali2005leveraging} present a multilingual SMT-based word alignment model, extending 
	IBM models, based on HMM models and a two step alignment procedure. Since the goal of 
	this research is to tackle word alignment  directly without considering 
	machine translation, these works are not considered here.

	In another line of research,  \citet{lardilleux:hal-00368737} introduce a corpus splitting 
	method to come up with a perfect alignment of multiwords.
	 \citet{lardilleux-lepage-2008-multilingual}, and \citet{lardilleux-lepage-2009-sampling}
	suggest to rely only on low frequency terms for a similar purpose: sub-sentential 
	alignment. These methods solve a somewhat different problem than what is addressed
	by us. Other usages of multiparallel corpora are language comparison
	 \cite{mayer-cysouw-2012-language}, typology
	studies  \cite{ostling-2015-word,
	asgari-schutze-2017-past,imanigooghari-etal-2021-parcoure} and SMT  \cite{nakov2012improving,
	 bertoldi2008phrase, dyer-etal-2013-simple}

\textbf{Matrix factorization and link prediction.} 
	Matrix factorization is a technique that factors, in
	the most typical case, a  matrix into two lower-ranked matrices
	in which the latent factors of the original matrix
	are represented. Matrix factorization approaches have been widely
	used in   document clustering  \cite{xu2003document, shahnaz2006document},
	topic modeling  \cite{kuang2015nonnegative,
	choo2013utopian} information
	retrieval  \cite{zamani2016pseudo,
	deerwester1990indexing} and NLP tasks like word
	sense disambiguation \cite{schutze-1998-automatic}.
	In 2009, Netflix's  recommender system competition revealed that this technique effectively works for 
	collaborative
	filtering   \cite{koren2009matrix}. Since then it has
	been a state of the art method in recommender systems.

	Link prediction algorithms are widely used in different areas of science since many social, biological,
	and information systems can be described as networks with nodes and connecting links  \cite{zhou2009predicting}. 
	Link prediction algorithms compute the likelihood of links based on different heuristics. 
	One can categorize available methods based on the maximum number of hops they consider in their
	computations for each node  \cite{zhang2018link}. First order algorithms, such as common neighbors (CN), only consider one hop neighborhoods,
	e.g.,  \cite{barabasi1999emergence}. Second order methods consider two hops, e.g.,   \cite{zhou2009predicting}.
	Finally, higher order methods take the whole network
	into account for making predictions
	 \cite{brin1998anatomy, jeh2002simrank,rothe-schutze-2014-cosimrank}.
In this paper, we use a two-hop method since it offers a
	good tradeoff between effectiveness and efficiency.

\section{Methods} 
\subsection{The \framework{} framework}
While a bilingual aligner considers each language pair separately, \framework{}  utilizes
the synergy between all language pairs to improve word alignment performance. In Figure \ref{fig:lexdiff},
Eflomal alignments of a sentence from PBC in four different languages are depicted. 
Although Eflomal has failed to find the link between German ``Schritt''
and French ``pas'', we can easily find this relation
by observing that the four nodes ``step'', ``Schritt'',
``paso'', and ``pas'' are fully connected, except for the
edge from ``Schritt'' to ``pas''. In this case, the
inference amounts to a completion of a clique.
However, most cases are not that simple. In the figure, English ``thousand miles''
is mistakenly aligned to Spanish ``siempre'' although its alignments to German ``lange'' and French ``mille'' are correct.
We would like to infer that ``thousand miles'' should be aligned
to ``largo'', but in this case creating a fully connected
subgraph, i.e., a clique (which would include ``siempre''),
would add many incorrect edges.
Given the complexity and error-proneness of
initial bilingual alignments,
inferring an alignment between two languages from a
multiparallel alignment in general is a complex problem.

Starting from a multiparallel corpus,
we first generate bilingual alignments for all language pairs.
\framework{} then employs a prediction algorithm to find and
add new alignment links.
In this paper, we focus on two prediction algorithms:
non-negative matrix factorization and Adamic-Adar link prediction.

\subsection{Non-negative matrix factorization}
Non-negative matrix factorization (NMF) has been used in
many different applications.
After discovery of its effectiveness for collaborative recommendation  \cite{koren2009matrix},
it was widely accepted as a standard method for recommender systems. 

In a standard recommender system  with \(m\) users
and \(n\) items, ratings (a number from 1 to 5) from each
user for the items they have seen so far are known. The aim
is to predict the ratings the user would give to unseen
items and, based on these predictions, recommend new items to
the user.  As described by \cite{luo2014efficient}, let
$W = \left[
w_{u,i}\right] \in \mathbb{R}^{m\times n}$ be
the matrix of ratings.
For
NMF to work it is essential that the matrix be sparse, thus
if a user's rating for an item is unknown, the corresponding
cell is zeroed.
The matrix \(W\) is then decomposed into
two low-rank non-negative matrices, $ T = \left[
t_{u,k} \right] \in \mathbb{R}^{m \times r} $ and \(V
= \left[ v_{k,i}\right] \in \mathbb{R}^{r \times n}\) such
that $TV \approx W$ and $ r \ll \min(m , n)$. $r$ is a
hyperparameter. By multiplication of these two matrices we
end up with a reduced matrix $ {W}' = TV $ in which
each 
zeroed cell $w_{u,i}$
from matrix $ W $
is replaced with a value $w_{u,i}'$
that represents a
prediction for the rating that user $u$ would give to item $i$.
NMF solves the following optimization program:
 \begin{eqnarray*}
 \underset{T,V}{\mbox{argmin}} \left( \| W - TV \|^2 \right)\\
 \mbox{subject to} \ \ \  T,V \geqslant 0
 \end{eqnarray*}

 This optimization problem can be 
 solved by gradient descent  using the following updates:
\begin{align*}
t_{u,k} \leftarrow t_{u,k} + \eta_{u,k} ((WV^T)_{u,k} - (TVV^T)_{u,k}) \nonumber\\
v_{k,i} \leftarrow v_{k,i} + \eta_{k,i} ((T^TW)_{k,i} - (T^TTV)_{k,i})
\end{align*}

In this equation, $ \eta $ is the learning rate. To
guarantee non-negativity, it
is defined as:
\begin{align*}
	\eta_{u,k} = \frac{t_{u,k}}{(TVV^T)_{u,k}},
	\ \ \eta_{k,i} = \frac{v_{k,i}}{(T^TTV)_{k,i}} 
\end{align*}

Note that the objective function only takes account of  non-zero cells.
 \citet{luo2014efficient} propose an approach that takes advantage of  the sparseness 
of the matrix for faster computation. In addition, Tikhonov regularization is integrated
to improve precision, recall, and convergence rate.

We use the implementaion of NMF provided by the
Surprise\footnote{\url{http://surpriselib.com/}} library,
with default
hyperparameters ($ r = 15$, $\mbox{n\_epochs} = 50$).

\subsubsection{NMF in \framework{} framework}

\def\mysep{0.3cm}

\begin{figure}[t]
	\centering
{\footnotesize
\begin{tabular}{l|@{\hspace{\mysep}}l@{\hspace{0.2cm}}l@{\hspace{\mysep}}l@{\hspace{\mysep}}l@{\hspace{\mysep}}l@{\hspace{\mysep}}l@{\hspace{\mysep}}l@{\hspace{\mysep}}l@{\hspace{\mysep}}l@{\hspace{\mysep}}l@{\hspace{\mysep}}l}
&\rotatebox{90}{I} & \rotatebox{90}{can} & \rotatebox{90}{see} & \rotatebox{90}{ich} & \rotatebox{90}{kann} & \rotatebox{90}{es} & \rotatebox{90}{sehen} & \rotatebox{90}{je} & \rotatebox{90}{vois}\\\hline
 I & 5 &&1&5& &1&&5&1\\
 can & &5&1&&5&&1&&\\
 see&1&1&5&&1&&5&1&5\\
 ich &5&1&&5&&&1&5&1\\
 kann & 1&5&&1 &5 &&&&&\\
 es&&&&&&5&1&&&\\
 sehen & 1&&5&&1&&5&1\\
 je&5&1&&5&&1&&5&1\\
 vois&1&&5&1&&&5&1&5
 \end{tabular}
}
	\caption{ An example of how the original matrix is filled for
	a sentence in three languages. Zero entries are left
	blank for readability.}
\label{fig:matrix}
\end{figure}

We create a separate matrix $W$ for each sentence in the multiparallel
corpus.
Tokens in the sentence play the role of both users
and items, i.e.,  we consider each token both as a row and
as a column. Figure \ref{fig:matrix} shows an example
of a sentence in a toy English-German-French multiparallel corpus.  If two tokens
are aligned using the bilingual aligner, we fill the
corresponding cell with the highest rating ($5$). To give a
few negative examples to the algorithm, if a token \(x\)
from language \( L_1 \) is aligned to token \(y\) in
language \( L_2 \), we pick another random token $z$ from $L_2$
and fill the corresponding cell of \(x\) to \(z\) with the
lowest rating ($1$). We zero out all other  cells.
Next we apply the matrix factorization algorithm
to this matrix and then compute the reduced matrix $W'$
from the factors. Now we grab the predicted alignment scores between
source and target languages from $W'$. To
extract new alignment edges we apply the Argmax
algorithm  \cite{jalili-sabet-etal-2020-simalign}.
Argmax creates an alignment edge between word $w_i$ from language 
$ L_1 $ and word $w_j$ from language $L_2$ if 
among all words from $L_2$, $w_i$ has the highest alignment score
with $w_j$, and among all words from $L_1$, $w_j$ has the highest alignment score
with $w_i$.

\subsection{Link Prediction}
A multiparallel sentence annotated with bilingual word
alignments
can be considered to be a
graph with words from all translations  as
nodes and the word alignments as edges.  Link prediction
algorithms such as Common Neighbors (CN) and Adamic-Adar (AdAd)
estimate the likelihood of having an edge
between two nodes $x$ and $y$
in the graph based on the similarity of their neighborhoods.
The CN index  weights all  common neighbors equally.
In contrast, AdAd gives higher weight to neighbors with low
degrees because sharing a neighbor that in turn has few
neighbors is more significant.
Therefore, we use the AdAd index.
It is defined as:
\begin{equation}
	\mbox{AdAd}_{x,y}=\sum_{z\in\Gamma(x)\cap\Gamma(y)}{\frac{1}{\log |\Gamma(z)|}}
\end{equation}
where $\Gamma(x)$ is the neighborhood of $x$.

If we use a word aligner that produces a  score for each
alignment edge, we can use Weighted Adamic-Adar  \cite{lu2010link}:
\begin{equation}
	\mbox{WAdAd}_{x,y}=\sum_{z\in\Gamma(x)\cap\Gamma(y)}{\frac{w(x,z)+w(z,y)}{\log(1+S(z))}}
\end{equation}
where $w(x,z)$ is the similarity score of $x$ and $z$
generated by the aligner and $S(x)=\sum_{z\in\Gamma(x)}{w(x,z)}$.
For embedding-based aligners we use embedding similarity as
the score $w(x,z)$. If an aligner does not provide scores,
we set all weights to 1.0.

Given a scored word alignment, we create a multilingual word
alignment matrix $W$ for each sentence as shown in
Figure~\ref{fig:matrix}. Each cell contains 0 or 1 for
Adamic-Adar or the alignment score for Weighted Adamic-Adar.
We again apply Argmax to extract new alignment edges and
then add them to the original alignment.

\begin{table*}[t]
	\centering
	\begin{threeparttable}
		\centering
		\scriptsize
		\begin{tabular}{ll||rrrr}
			&Language Pair & Name & \# Sentences (train/valid./test) \\
			\midrule
			\midrule
			\multirow{4}{0.5cm}{Bible}&FIN-HEB & HELFI  \cite{yli-jyra-etal-2020-helfi} & 22291 (17832/2229/2230)\\
			&FIN-GRC & HELFI  \cite{yli-jyra-etal-2020-helfi} & 7909 (6327/791/791) \\
			&ENG-FRA & BLINKER  \cite{melamed1998blinker}  & 250\\
			\midrule
			\multirow{4}{0.5cm}{Non-Bible}&ENG-DEU & EuroParl-based\tnote{a}   &  508 \\
			&ENG-FAS &  \cite{tavakoli2014phrase} &  400 \\
			&ENG-HIN & WPT2005\tnote{b} & 90 \\
			&ENG-RON & WPT2005\tnote{b}& 203 \\
		\end{tabular}
		\begin{tablenotes}
			\item[a]  \url{www-i6.informatik.rwth-aachen.de/goldAlignment/} 
			\item[b] \url{http://web.eecs.umich.edu/~mihalcea/wpt05/}
		\end{tablenotes}
	\end{threeparttable}
	\caption{Overview of datasets. We use ISO 639-3 language codes. \# Sentences:
	the number of available verses (i.e., sentences). FIN-HEB and FIN-GRC
	datasets split into train, validation and test.
	\tablabel{data}}
\end{table*}

\section{Experimental setup}

\subsection{PBC}

The PBC corpus  \citep{mayer-cysouw-2014-creating} contains \neditions{}  editions
of the Bible in \nlangs{} languages. The editions are aligned at the verse level
and tokenized. A verse can contain more than one sentence,
but
we treat it as one unit in the parallel corpus since a true
sentence level alignment is not available.
There are some errors in  tokenization  (e.g., for Tibetan,
Khmer and Chinese), but the overall quality of the
corpus is good.
For the majority of languages one edition is provided, while a few languages
(in particular, English, French and German) contain up to
dozens of editions. The verse coverage also 
differs from  language to language. Some languages have
translations of
both New Testament and Hebrew Bible while others contain
only one.
Table \ref{tab:pbc_statics} gives
corpus statistics.

\begin{table}
	\centering \footnotesize 
	\begin{tabular}
		{lr} \toprule \# editions & \neditions{} \\
		\# languages & \nlangs{} \\
		\# verses & \nversesexact{} \\
		\# verses / \# editions & \avgverses{} \\
		\# tokens / \# verses & \avgtokens{} \\
		\bottomrule 
	\end{tabular}
	\caption{PBC corpus statistics\tablabel{stats}}
	 \label{tab:pbc_statics}
\end{table}

\subsection{Word alignment datasets}
PBC does not provide gold word  alignments. To evaluate \framework{}, we port two
word alignment gold datasets of the Bible to PBC:
Blinker  \cite{melamed1998blinker}  and the recently
published HELFI  \cite{yli-jyra-etal-2020-helfi}.
We further experiment with bilingual datasets, using
Machine Translation (MT) to create multiparallel corpora.
Table \ref{tab:data} gives dataset statistics.

\textbf{The HELFI dataset} consists of
the Greek New Testament, the 
Hebrew Bible and
translations of both into 
Finnish. In addition,
morpheme alignments are provided for 
Finnish-Greek and Finnish-Hebrew. We reformatted this dataset to the format used by 
PBC. In more detail, we added three new editions for the three languages to PBC. 
We identified the PBC verse identifier for each verse of
HELFI to ensure proper verse alignment of these three new editions.
 The Finnish-Hebrew dataset has
22,291 verses and the Finnish-Greek dataset 7,909.
We split these datasets 80/10/10 into  train, validation and test.

\textbf{The Blinker Bible dataset} provides word level alignments of 250 
Bible verses between English and French. The French side of this dataset matches
with
the edition Louis Segond 1910 in PBC. However, the
tokenizations (Blinker vs PBC) are different. 
We therefore create a mapping of the tokens using character n-gram matching.
For English, we created and added a 
new edition to PBC. 

\textbf{MT datasets}. 
To more broadly evaluate \framework{},
we also create
multiparallel datasets for four non-Bible
word alignment gold standards; these are listed
in \tabref{data} as ``Non-Bible'' corpora. 
For these gold standards, we translate 
from English to all languages available in  Google
Translate, using their
API.\footnote{\url{https://cloud.google.com/translate/docs/basic/translating-text}} For
the added languages, we create alignments for the gold
standard sentences using
\simalign{}.

\subsection{Initial word alignments} 
\label{experiment_setup_init_word_aligns}
We compare with two state of the art models, one statistical, one neural.
Eflomal  \cite{ostling2016efficient} is a Bayesian statistical
word aligner using
Markov Chain Monte Carlo inference. \simalign{}  \cite{jalili-sabet-etal-2020-simalign}
obtains word alignments from multilingual pretrained language models with no need for parallel data.
For the symmetrization of Eflomal, we use grow-diag-final-and (GDFA)
and intersection, and for \simalign{} we use Argmax and Itermax.
Intersection and Argmax generate accurate alignments
while GDFA and Itermax are less accurate but have better coverage \cite{jalili-sabet-etal-2020-simalign}.

We evaluate on a \emph{target language
pair} parallel sentence as follows:
First, we create the matrix (Figure \ref{fig:matrix}) for
this sentence for all languages in the multiparallel corpus.
Then we run link prediction on the matrix -- this
accumulates evidence from a set of languages in the multiparallel corpus.
Finally, we take the predictions for the target language pair
and add them to the original (bilingual) alignment.

NMF works best if it starts with
high-accuracy (i.e., non-noisy) bilingual alignments -- 
errors can result in incorrectly predicted alignment edges.
We therefore use SimAlign Argmax and Eflomal Intersection,
two word alignment methods with high precision, to create
the initial alignments that are then fed into NMF.
We then add the predictions to any desired original alignments; e.g., NMF (GDFA) uses Eflomal Intersection as the initial alignments and adds the predictions to Eflomal GDFA.
See the Appendix for more details.

\simalign{} offers high quality word alignments for well-represented 
languages from pretrained language models; 
however, our experiments show that its performance is 
far behind Eflomal for less well resourced languages 
like Biblical Hebrew and Koine Greek. 
Also, Eflomal is a better match for
\framework{} because it can provide word alignments for all languages available
in a multiparallel corpus. In contrast,  \simalign{}
is limited to languages supported by pretrained multilingual embeddings.

To feed Eflomal with enough training data for a target language pair,
we use all available data from different translations of the language pair.
For example if one language has two translations and the other one has three translations,
Eflomal's training data will contain six aligned translation pairs for these
two languages.

We use the standard evaluation measures for word alignment:
precision, recall, $F_1$ and Alignment Error Rate (AER)
 \cite{och-ney-2003-systematic,ostling2016efficient,jalili-sabet-etal-2020-simalign}.

\section{Results} 

\subsection{Multiparallel corpus results} 
We perform the first set of experiments on the Blinker Bible and the HELFI gold standards in the PBC.
The baseline results are calculated on the original language pairs. \framework{} can be applied to both Eflomal and SimAlign alignments.
Since the default version of SimAlign can only generate alignments for the 84 languages that 
multilingual BERT supports,\footnote{\url{https://github.com/google-research/bert/blob/master/multilingual.md}}
for a better comparison, we use the same set of languages in the alignment graph for both SimAlign and Eflomal.

\begin{table*}[t]
	\scriptsize
	\centering
	\def\tablesep{0.1cm}
	\begin{tabular}{
			@{\hspace{\tablesep}}c@{\hspace{\tablesep}}
			@{\hspace{\tablesep}}c@{\hspace{\tablesep}}|
			@{\hspace{\tablesep}}l@{\hspace{\tablesep}}||
			@{\hspace{\tablesep}}c@{\hspace{\tablesep}}
			@{\hspace{\tablesep}}c@{\hspace{\tablesep}}
			@{\hspace{\tablesep}}c@{\hspace{\tablesep}}
			@{\hspace{\tablesep}}c@{\hspace{\tablesep}}|
			@{\hspace{\tablesep}}c@{\hspace{\tablesep}}
			@{\hspace{\tablesep}}c@{\hspace{\tablesep}}
			@{\hspace{\tablesep}}c@{\hspace{\tablesep}}
			@{\hspace{\tablesep}}c@{\hspace{\tablesep}}|
			@{\hspace{\tablesep}}c@{\hspace{\tablesep}}
			@{\hspace{\tablesep}}c@{\hspace{\tablesep}}
			@{\hspace{\tablesep}}c@{\hspace{\tablesep}}
			@{\hspace{\tablesep}}c@{\hspace{\tablesep}}
			}
		& & & \multicolumn{4}{c}{ FIN-HEB } & \multicolumn{4}{c}{ FIN-GRC } & \multicolumn{4}{c}{ ENG-FRA} \\
		& & Method & Prec. & Rec. & $F_1$ & AER & Prec. & Rec. & $F_1$ & AER & Prec. & Rec. & $F_1$  & AER \\
		\midrule
		\midrule
		\multirow{4}{*}{ \rotatebox{90}{ } } & \multirow{4}{*}{ Baseline }  
		  & Eflomal (intersection) & \bfseries0.818 & 0.269          & 0.405           & 0.595          & \bfseries0.897 & 0.506          & 0.647          & 0.353          & \bfseries0.971 & 0.521          & 0.678          & 0.261         \\
		& & Eflomal (GDFA)         & 0.508          & 0.448          & 0.476           & 0.524          & 0.733          & 0.671          & 0.701          & 0.300          & 0.856          & 0.710          & 0.776          & 0.221         \\
		& & SimAlign      & 0.190          & 0.113          & 0.142           & 0.858          & 0.366          & 0.265          & 0.307          & 0.693          & 0.886          & 0.692          & 0.777          & 0.221         \\
		\midrule
		\midrule
		\multirow{3}{*}{ \rotatebox{90}{ } }& \multirow{3}{*}{ Init  SimAlign }
		& AdAd            & 0.199          & 0.127          & 0.155           & 0.845          & 0.402          & 0.289          & 0.336          & 0.664          & 0.878          & 0.731          & 0.798          & 0.200       \\
		& & WAdAd           & 0.186          & 0.165          & 0.175           & 0.825          & 0.353          & 0.350          & 0.351          & 0.649          & 0.856          & 0.752          & 0.801          & 0.197       \\
		& & NMF           & 0.122          & 0.100          & 0.110           & 0.890          & 0.396          & 0.337          & 0.364          & 0.636          & 0.835          & 0.700          & 0.762          & 0.236         \\
		\midrule
		\multirow{4}{*}{ \rotatebox{90}{ } }& \multirow{4}{*}{ Init  Eflomal }  
		& WAdAd (intersection)          & 0.781          & 0.612          & \bfseries0.686  & \bfseries0.314 & 0.849          & 0.696          & \bfseries0.765 & \bfseries0.235 & 0.938          & 0.689          & 0.794          & 0.203           \\
		& & NMF (intersection)  & 0.78 & 0.576 & 0.663 & 0.337 &    0.864 & 0.669 & 0.754 & 0.248 &    0.948 & 0.624 & 0.753 & 0.245 \\
		\cmidrule{3-15}
		& & WAdAd (GDFA)            & 0.546          & \bfseries0.693 & 0.611           & 0.389          & 0.707          & \bfseries0.783 & 0.743          & 0.257          & 0.831          & \bfseries0.796 & \bfseries0.813 & \bfseries0.186 \\
		& & NMF (GDFA)                    & 0.548          & 0.646 & 0.593           & 0.407          & 0.72          & 0.759 & 0.739          & 0.261          & 0.844          & 0.767 & 0.804 & 0.195  \\
\midrule\midrule
	\end{tabular}
	\caption{Comparison of results from different methods on PBC.
	The best results in each column are in bold. The
	three methods exploiting multiparallelism (AdAd,
	WAdAd, NMF) generally outperform the baselines on
	$F_1$ and AER.}
	\tablabel{res_pbc}
\end{table*}

Table \ref{tab:res_pbc} shows the results for our methods
applied on SimAlign and Eflomal baselines.\footnote{We only
consider SimAlign IterMax, not SimAlign ArgMax, because
IterMax performed better throughout.}  AdAd, NMF and
WAdAd substantially improve the performance for all
language pairs.  SimAlign  generates high-quality
alignments for the English-French dataset, but cannot properly
align underresourced languages like Biblical Hebrew and
Koine Greek.
In such cases, \framework{} uses the accumulated information
from all other language pairs in
the graph to improve the
performance.  When starting with the SimAlign alignment
(``Init SimAlign''), both methods improve the  result for
both FIN-HEB and FIN-GRC.

Eflomal generates better alignments for FIN-HEB and FIN-GRC.
This means that Eflomal also generates better alignments
between FIN, HEB and GRC on the one hand and the other
languages 
in the graph on the other hand and therefore can provide a better signal for \framework{}.
The improvements of our models applied on Eflomal are higher than the ones applied on SimAlign for these language pairs.

When changing the initial alignments from Eflomal (intersection) to Eflomal (GDFA), we see different behaviors:
GDFA improves the results for Blinker  while it does not help for HELFI.
We believe this is caused by the different ways the two
datasets were annotated.
In Blinker, many phrases are ``exhaustively'' aligned:
if a phrase DE in English is aligned with FG in French then
all four alignment edges (D-F, D-G, E-F, E-G) are given as
gold edges.\footnote{For example,
the alignment of the phrases ``trembled violently'' and
``fut saisi d'und grande, d'une violente \'{e}motion''
consists of $2 \cdot 8$
gold edges.}

So Blinker contains a lot of
many-to-many links.
In contrast, most alignments are one-to-one in HELFI.
This partially explains why intersection as initial
alignment works much better for HELFI than GDFA and vice
versa for Blinker.

In summary, compared to the baselines, we see very large improvements through exploiting
multiparallelism for one type of alignment methodology
(HELFI, $F_1$ improved by up to 20\% for FIN-HEB) and improvements of up
to 3.5\% for the other (ENG-FRA).

\subsection{MT dataset results}
We perform the second set of experiments on gold standard alignments for language pairs that are not part of a multiparallel corpus such as PBC.
To this end, we create artificial multiparallel corpora by translating the English
side to all languages available in the Google Translate API.
The main goal is to give broader evidence for the
effectiveness of our method, beyond the specialized domain
of the Bible.

Eflomal's alignments generally have good quality. However, 
they
get worse when less parallel data is available \cite{jalili-sabet-etal-2020-simalign}.
Since the size of the
multiparallel corpus created by machine translation
is rather small, we use SimAlign for generating initial
alignments.
SimAlign has been shown to have good performance even for
very small parallel corpora; in fact, it does not 
need any parallel data at all.

Table \ref{tab:res_mt} shows the results of the experiments.
Both NMF and WAdAd,
improve the performance of the baseline by using the
alignment graph. Improvements range from 0.8\% (ENG-DEU)  to 3.3\% (ENG-HIN).
This again demonstrates the utility of exploiting multiparallelism
for word alignment.
It is worth mentioning that
in this case the translations are noisy since they were
automatically generated.
But even with these
noisy translations (instead of a ``true'' multiparallel
corpus), our models  effectively leverage
multiparallelism.

\begin{table*}[t]
	\scriptsize
	\centering
	\def\tablesep{0.1cm}
	\begin{tabular}{
			@{\hspace{\tablesep}}c@{\hspace{\tablesep}}|
			@{\hspace{\tablesep}}l@{\hspace{\tablesep}}||
			@{\hspace{\tablesep}}c@{\hspace{\tablesep}}
			@{\hspace{\tablesep}}c@{\hspace{\tablesep}}
			@{\hspace{\tablesep}}c@{\hspace{\tablesep}}
			@{\hspace{\tablesep}}c@{\hspace{\tablesep}}|
			@{\hspace{\tablesep}}c@{\hspace{\tablesep}}
			@{\hspace{\tablesep}}c@{\hspace{\tablesep}}
			@{\hspace{\tablesep}}c@{\hspace{\tablesep}}
			@{\hspace{\tablesep}}c@{\hspace{\tablesep}}|
			@{\hspace{\tablesep}}c@{\hspace{\tablesep}}
			@{\hspace{\tablesep}}c@{\hspace{\tablesep}}
			@{\hspace{\tablesep}}c@{\hspace{\tablesep}}
			@{\hspace{\tablesep}}c@{\hspace{\tablesep}}|
			@{\hspace{\tablesep}}c@{\hspace{\tablesep}}
			@{\hspace{\tablesep}}c@{\hspace{\tablesep}}
			@{\hspace{\tablesep}}c@{\hspace{\tablesep}}
			@{\hspace{\tablesep}}c@{\hspace{\tablesep}}}
		& & \multicolumn{4}{c}{ ENG-PES } & \multicolumn{4}{c}{ ENG-HIN } & \multicolumn{4}{c}{ ENG-RON } & \multicolumn{4}{c}{ ENG-DEU } \\
		& Method & Prec. & Rec. & $F_1$ & AER & Prec. & Rec. & $F_1$ & AER & Prec. & Rec. & $F_1$ & AER & Prec. & Rec. & $F_1$ & AER \\
		\midrule
		\midrule
		 Baseline   
		  & SimAlign      & \bfseries0.756          & 0.645         & 0.696           & 0.304            & \bfseries0.709          & 0.493          & 0.582          & 0.418          & \bfseries0.807          & 0.663          & 0.728          & 0.272          & \bfseries0.829           & 0.795          & 0.812          & 0.188 \\
		\midrule                                               
		\midrule                                              
		\multirow{3}{*}{ Init  SimAlign }                                                 
		& AdAd            & 0.751          & 0.700          & \bfseries0.725          & \bfseries0.276          & 0.693          & 0.544          & 0.610          & 0.390          & 0.799          & 0.696          & \bfseries0.744 & \bfseries0.256 & 0.818           & 0.823          & \bfseries0.820 & \bfseries0.179 \\
		& WAdAd          & 0.705          & \bfseries0.740 & 0.722          & 0.278          & 0.643          & \bfseries0.574 & 0.607          & 0.394          & 0.725          & \bfseries0.717 & 0.721          & 0.279          & 0.749           & \bfseries0.844 & 0.794          & 0.207 \\
		& NMF  & 0.734 & 0.698 & 0.716 & 0.284 &    0.684 & 0.559 & \bfseries0.615 & \bfseries0.385 &    0.780 & 0.696 & 0.736 & 0.265 &    0.804 & 0.827 & 0.815 & 0.185 \\
		\midrule
\midrule
	\end{tabular}
	\caption{Results with gold standards translated into
	other languages using machine translation.
	The best results in each column are in bold.
        The
	three methods exploiting multiparallelism (AdAd,
	WAdAd, NMF) outperform the baselines on
	$F_1$ and AER.}
	\tablabel{translate}
	\label{tab:res_mt}
\end{table*}

\subsection{Analysis} 

\subsubsection{Effect of number of languages}  
\begin{figure}[t]
	\centering
	\includegraphics[width=1.0\linewidth]{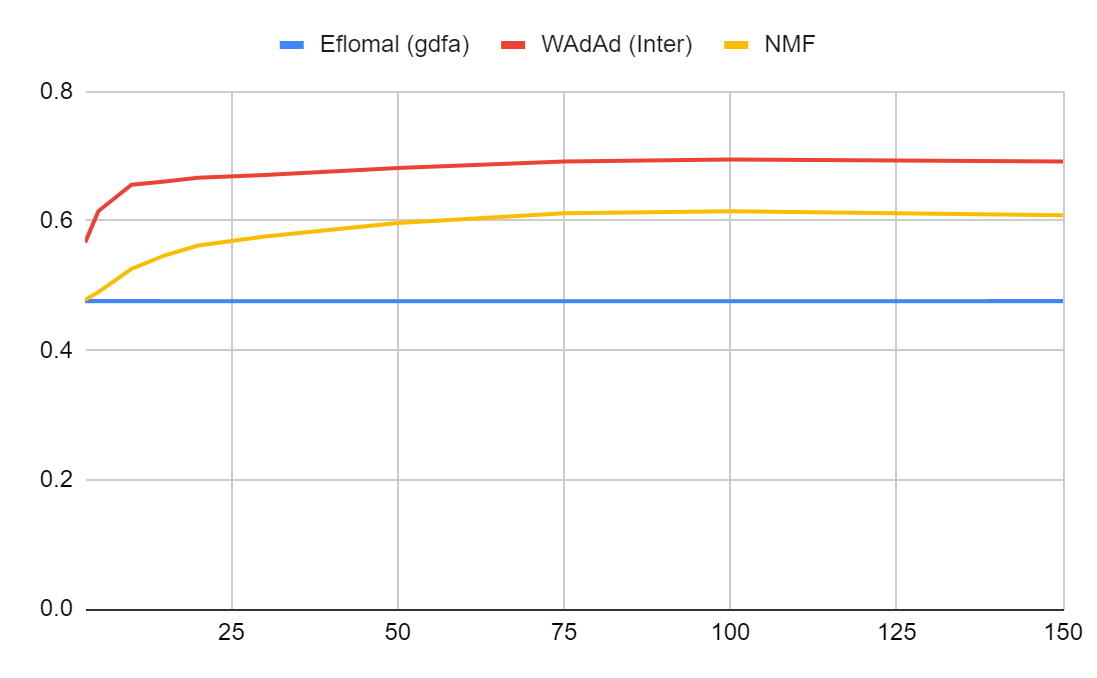}
	\caption{$F_1$ of \framework{} for the target language
	pair FIN-HEB as a function of the number 
	of additional languages. There is a clear rise
	initially. The curve flattens around 75.}
\label{fig:lang_count}
\end{figure}

The effect of adding more languages to the alignment graph is depicted in Figure \ref{fig:lang_count}.
This plot shows $F_1$ for  FIN-HEB.
As seen in the figure, the slope is pretty steep up to 25 languages, but even adding just three languages can still improve the results.
For 75 languages we have almost reached the peak
and after 100, adding more languages is not improving the
results.
This means that \framework{} can also be helpful for corpora with a
smaller number of
languages -- a massively parallel corpus with 
thousands of languages is not required.

\subsubsection{Size of the training set}
To assess the effect of dataset size on the performance
of \framework, 
we perform a set of experiments on ENG-FRA and NMF.
To this end, we take the training data for ENG-FRA and train
models on subsets of it. The training data consists of 6.4M
sentence pairs -- this number is so high because 
we use the crossproduct of all editions in English and French
(\S \ref{experiment_setup_init_word_aligns}).

\begin{figure}[t]
	\centering
	\includegraphics[width=0.95\linewidth]{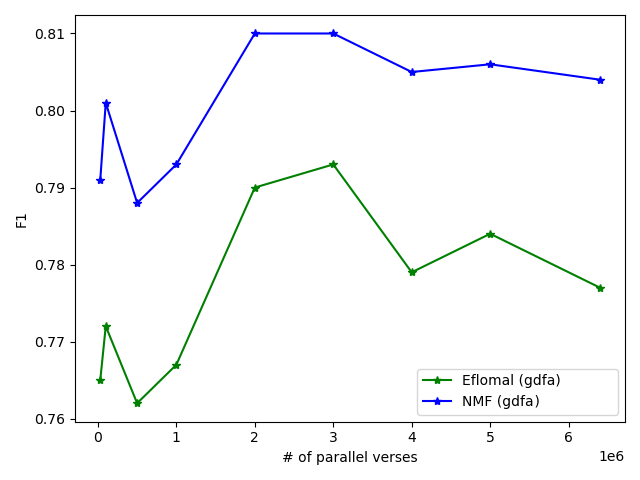}
	\caption{Word alignment $F_1$ on ENG-FRA as a function of
	the size of the training set, ranging from 30K to
	6.4M training sentence pairs
	}
\label{fig:training_size}
\end{figure}

The results are shown in Figure~\ref{fig:training_size}.
Eflomal performance  increases with training set
size initially and is then less predictable.
NMF consistently improves the scores.

\subsubsection{Effect of task difficulty}
\label{effectdifficulty}
Table \ref{tab:res_pbc} shows large improvements for all datasets, and especially
for FIN-HEB and FIN-GRC.
To get more insight into the reasons for this improvement,
we stratify
 FIN-HEB verses by dividing the interval $[0,1]$ of 
initial $F_1$ performance of
Eflomal into five equal-sized subintervals: $[0,0.2]$, \ldots, $(0.8,1]$.

\begin{figure}[t]
	\centering
	\includegraphics[width=1.0\linewidth]{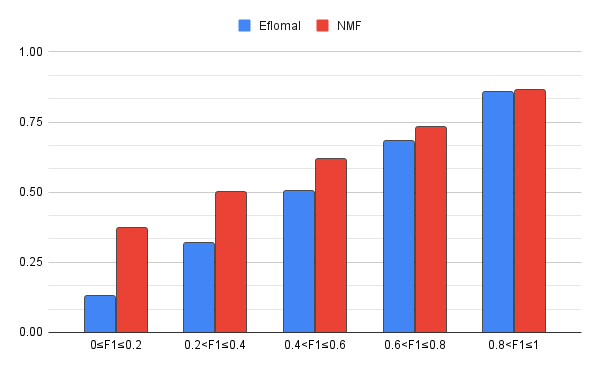}
	\caption{How helpful is \framework{} for different
	difficulty levels? For this analysis, FIN-HEB verses were
	stratified according to Eflomal $F_1$ (x-axis).
	We see that \framework{} brings the largest
	improvements for difficult sentences.
	}
\label{fig:difficulty_level}
\end{figure}

Figure \ref{fig:difficulty_level} indicates that
\framework{} is most effective for  difficult verses, but
brings little  improvement 
for easy verses. We attribute this to two reasons: 
\begin{enumerate}
	\item An easy to align verse in a language pair cannot use help from
	other languages since it already has good alignment
	links (although
	the language pair would still be of benefit in
	improving alignments for the sentence in other languages).
 So there is no way for \framework{} to get better
	results in this scenario.
	\item \framework{} only tries to get better results by adding new alignments, 
	and as an easy verse already has many alignment links, adding new links 
almost inevitably results in a drop in precision.
It may also be possible to inspect and
	prune existing Eflomal links 
	using \framework{} to get better results in this scenario.
\end{enumerate} 

\subsubsection{Most helpful languages}
For each dataset, the five most helpful languages with their
corresponding improvements are listed in Table \ref{p:tab:helpful_langs}.
We hypothesize that these languages serve to
bridge the typological gap between the two target
languages.  Table \ref{p:tab:helpful_langs} suggests one
should be able to achieve excellent results -- even for a
corpus with a small number of languages -- if we utilize an
intelligent selection of languages.

\begin{table}
	\scriptsize
	\centering
	\def\tablesep{0.1cm}
	\begin{tabular}{cr|cr|cr}
		\multicolumn{2}{c}{ENG-FRA} & \multicolumn{2}{c}{FIN-HEB} & \multicolumn{2}{c}{FIN-GRC} \\
		Lang. & $\Delta$ & Lang. & $\Delta$ & Lang. & $\Delta$ \\
		\midrule
		SPA & +2.0\% & TGL & +17.7\% & LAT & +7.5\% \\
		ITA & +1.9\% & FRY,ELL & +17.3\% & ELL & +6.6\% \\
		DEU & +1.8\% & SWE & +17.3\% & ENG & +6.1\% \\
		NLD & +1.4\% & NLD & +16.8\% & FRY & +5.8\% \\
		AFR & +1.3\% & YOR & +14.2\% & BEL & +5.7\% \\
	\end{tabular}
	\caption{The five most helpful languages and WAdAd's
	absolute improvements in $F_1$
 over the initial bilingual aligner SimAlign. For example,
\framework{} improves the bilingual FIN-GRC alignment by 7.5\% if
	applied to the trilingual corpus FIN-GRC-LAT, i.e.,
        Latin can be viewed as the best bridge between
        Finnish and Greek.}
	\tablabel{tab:helpful_langs}
\end{table}

\subsubsection{Multiple translations in two languages}
There are some datasets that contain few languages, but many
translations of a text in one language. PBC is one example
of such a dataset,
many literary works another (e.g., many novels have many translations in
English).
To see whether \framework{} can also help in this scenario, 
we picked all available 49 English and French editions from PBC and used them
as additional translations for the ENG-FRA dataset. The outcome of this
experiment is compared with the outcome of the same setup,
but with translations from languages other than French and English
in Table \ref{p:tab:two_lang_translations}. 
From this table we can conclude that translations from the 
target language pair can also assist, but not as much as 
translations from other languages. 
\begin{table}
	\scriptsize
	\centering
	\def\tablesep{0.1cm}
	\begin{tabular}{ccccc}
		& Prec. & Rec. & $F_1$ & AER\\
		\midrule
		Eflomal (intersection)         &0.971   & 0.521  & 0.678 & 0.319 \\
		Eflomal (GDFA)                 &0.856   & 0.710  & 0.776 & 0.221 \\
		\midrule
		NMF (target languages)         & 0.830 & 0.749 & 0.787  & 0.213 \\
		NMF (other languages)   & 0.837 & 0.753 & 0.793  & 0.205 \\
	\end{tabular}

	\caption{$F_1$ for ENG-FRA. \framework{} can
	  exploit a multiparallel corpus with languages
	  different from the target languages (``other
	  languages'') better than one that contains
	  additional translations in the target languages
	  (``target languages'').}
	\tablabel{tab:two_lang_translations}
\end{table}

\section{Conclusion and Future Work}
We presented \framework{}, a framework for leveraging
multiparallel corpora for word alignment.  We used two
prediction methods, one based on recommender systems and one
based on link prediction algorithms.  By adding new
alignment edges to the output of bilingual aligners, both
methods show large improvements over the bilingual baselines, with
absolute improvements of $F_1$
of up to 20\%.
We have also
ported Blinker and HELFI word alignment gold
standards to the Parallel Bible Corpus in the hope
that this will help foster
more work on exploiting multiparallel corproa.

\textbf{Future work.} In this paper, we have mainly focused on \emph{adding} new
alignment edges to baseline word alignments based on
properties of the multiparallel alignment graph. This
increases recall, but can harm precision. In future work, we
plan to expand on the possibility of \emph{deleting} edges  based on evidence from the
multiparallel alignment graph (cf.\ \ref{effectdifficulty}), thereby potentially improving
both precision and recall.

\section*{Acknowledgments} 
This work was supported by the European Research Council
(ERC, Grant No.\ 740516) and the German Federal Ministry of Education and
Research (BMBF, Grant No.\ 01IS18036A). The fourth author
was supported by the Bavarian research institute for digital
transformation (bidt) through their fellowship program. We
thank the anonymous reviewers for their constructive
comments.

\bibliographystyle{acl_natbib} 
\bibliography{anthology,acl2021}

\begin{thebibliography}{57}
\expandafter\ifx\csname natexlab\endcsname\relax\def\natexlab#1{#1}\fi

\bibitem[{Adamic and Adar(2003)}]{adamic2003friends}
Lada~A Adamic and Eytan Adar. 2003.
\newblock \href
  {https://citeseerx.ist.psu.edu/viewdoc/download?doi=10.1.1.108.1370&rep=rep1&type=pdf}
  {Friends and neighbors on the web}.
\newblock \emph{Social networks}, 25(3):211--230.

\bibitem[{Alkhouli et~al.(2016)Alkhouli, Bretschner, Peter, Hethnawi, Guta, and
  Ney}]{alkhouli-etal-2016-alignment}
Tamer Alkhouli, Gabriel Bretschner, Jan-Thorsten Peter, Mohammed Hethnawi,
  Andreas Guta, and Hermann Ney. 2016.
\newblock \href {https://doi.org/10.18653/v1/W16-2206} {Alignment-based neural
  machine translation}.
\newblock In \emph{Proceedings of the First Conference on Machine Translation:
  Volume 1, Research Papers}, pages 54--65, Berlin, Germany. Association for
  Computational Linguistics.

\bibitem[{Alkhouli and Ney(2017)}]{alkhouli-ney-2017-biasing}
Tamer Alkhouli and Hermann Ney. 2017.
\newblock \href {https://doi.org/10.18653/v1/W17-4711} {Biasing attention-based
  recurrent neural networks using external alignment information}.
\newblock In \emph{Proceedings of the Second Conference on Machine
  Translation}, pages 108--117, Copenhagen, Denmark. Association for
  Computational Linguistics.

\bibitem[{Asgari and Sch{\"u}tze(2017)}]{asgari-schutze-2017-past}
Ehsaneddin Asgari and Hinrich Sch{\"u}tze. 2017.
\newblock \href {https://doi.org/10.18653/v1/D17-1011} {Past, present, future:
  A computational investigation of the typology of tense in 1000 languages}.
\newblock In \emph{Proceedings of the 2017 Conference on Empirical Methods in
  Natural Language Processing}, pages 113--124, Copenhagen, Denmark.
  Association for Computational Linguistics.

\bibitem[{Barab{\'a}si and Albert(1999)}]{barabasi1999emergence}
Albert-L{\'a}szl{\'o} Barab{\'a}si and R{\'e}ka Albert. 1999.
\newblock \href {https://doi.org/10.1126/science.286.5439.509} {Emergence of
  scaling in random networks}.
\newblock \emph{science}, 286(5439):509--512.

\bibitem[{Bertoldi et~al.(2008)Bertoldi, Barbaiani, Federico, and
  Cattoni}]{bertoldi2008phrase}
Nicola Bertoldi, Madalina Barbaiani, Marcello Federico, and Roldano Cattoni.
  2008.
\newblock \href
  {http://citeseerx.ist.psu.edu/viewdoc/download?doi=10.1.1.528.5311&rep=rep1&type=pdf}
  {Phrase-based statistical machine translation with pivot languages}.
\newblock In \emph{International Workshop on Spoken Language Translation
  (IWSLT) 2008}.

\bibitem[{Brin and Page(1998)}]{brin1998anatomy}
Sergey Brin and Lawrence Page. 1998.
\newblock \href {https://doi.org/10.1016/S0169-7552(98)00110-X} {The anatomy of
  a large-scale hypertextual web search engine}.
\newblock \emph{Computer networks and ISDN systems}, 30(1-7):107--117.

\bibitem[{Brown et~al.(1993)Brown, Della~Pietra, Della~Pietra, and
  Mercer}]{brown1993mathematics}
Peter~F. Brown, Stephen~A. Della~Pietra, Vincent~J. Della~Pietra, and Robert~L.
  Mercer. 1993.
\newblock \href {https://www.aclweb.org/anthology/J93-2003} {The mathematics of
  statistical machine translation: Parameter estimation}.
\newblock \emph{Computational Linguistics}, 19(2).

\bibitem[{Choo et~al.(2013)Choo, Lee, Reddy, and Park}]{choo2013utopian}
Jaegul Choo, Changhyun Lee, Chandan~K Reddy, and Haesun Park. 2013.
\newblock \href {https://doi.org/10.1109/TVCG.2013.212} {Utopian: User-driven
  topic modeling based on interactive nonnegative matrix factorization}.
\newblock \emph{IEEE transactions on visualization and computer graphics},
  19(12):1992--2001.

\bibitem[{Cohn and Lapata(2007)}]{cohn-lapata-2007-machine}
Trevor Cohn and Mirella Lapata. 2007.
\newblock \href {https://www.aclweb.org/anthology/P07-1092} {Machine
  translation by triangulation: Making effective use of multi-parallel
  corpora}.
\newblock In \emph{Proceedings of the 45th Annual Meeting of the Association of
  Computational Linguistics}, pages 728--735, Prague, Czech Republic.
  Association for Computational Linguistics.

\bibitem[{Conneau et~al.(2020)Conneau, Khandelwal, Goyal, Chaudhary, Wenzek,
  Guzm{\'a}n, Grave, Ott, Zettlemoyer, and
  Stoyanov}]{conneau-etal-2020-unsupervised}
Alexis Conneau, Kartikay Khandelwal, Naman Goyal, Vishrav Chaudhary, Guillaume
  Wenzek, Francisco Guzm{\'a}n, Edouard Grave, Myle Ott, Luke Zettlemoyer, and
  Veselin Stoyanov. 2020.
\newblock \href {https://doi.org/10.18653/v1/2020.acl-main.747} {Unsupervised
  cross-lingual representation learning at scale}.
\newblock In \emph{Proceedings of the 58th Annual Meeting of the Association
  for Computational Linguistics}, pages 8440--8451, Online. Association for
  Computational Linguistics.

\bibitem[{Deerwester et~al.(1990)Deerwester, Dumais, Furnas, Landauer, and
  Harshman}]{deerwester1990indexing}
Scott Deerwester, Susan~T Dumais, George~W Furnas, Thomas~K Landauer, and
  Richard Harshman. 1990.
\newblock \href
  {https://doi.org/10.1002/(SICI)1097-4571(199009)41:6%3C391::AID-ASI1%3E3.0.CO;2-9}
  {Indexing by latent semantic analysis}.
\newblock \emph{Journal of the American society for information science},
  41(6):391--407.

\bibitem[{Devlin et~al.(2019)Devlin, Chang, Lee, and
  Toutanova}]{devlin-etal-2019-bert}
Jacob Devlin, Ming-Wei Chang, Kenton Lee, and Kristina Toutanova. 2019.
\newblock \href {https://doi.org/10.18653/v1/N19-1423} {{BERT}: Pre-training of
  deep bidirectional transformers for language understanding}.
\newblock In \emph{Proceedings of the 2019 Conference of the North {A}merican
  Chapter of the Association for Computational Linguistics: Human Language
  Technologies, Volume 1 (Long and Short Papers)}, pages 4171--4186,
  Minneapolis, Minnesota. Association for Computational Linguistics.

\bibitem[{Dou and Neubig(2021)}]{dou-neubig-2021-word}
Zi-Yi Dou and Graham Neubig. 2021.
\newblock \href {https://aclanthology.org/2021.eacl-main.181} {Word alignment
  by fine-tuning embeddings on parallel corpora}.
\newblock In \emph{Proceedings of the 16th Conference of the European Chapter
  of the Association for Computational Linguistics: Main Volume}, pages
  2112--2128, Online. Association for Computational Linguistics.

\bibitem[{Dufter et~al.(2018)Dufter, Zhao, Schmitt, Fraser, and
  Sch{\"u}tze}]{dufter-etal-2018-embedding}
Philipp Dufter, Mengjie Zhao, Martin Schmitt, Alexander Fraser, and Hinrich
  Sch{\"u}tze. 2018.
\newblock \href {https://doi.org/10.18653/v1/P18-1141} {Embedding learning
  through multilingual concept induction}.
\newblock In \emph{Proceedings of the 56th Annual Meeting of the Association
  for Computational Linguistics (Volume 1: Long Papers)}, pages 1520--1530,
  Melbourne, Australia. Association for Computational Linguistics.

\bibitem[{Dyer et~al.(2013)Dyer, Chahuneau, and Smith}]{dyer-etal-2013-simple}
Chris Dyer, Victor Chahuneau, and Noah~A. Smith. 2013.
\newblock \href {https://www.aclweb.org/anthology/N13-1073} {A simple, fast,
  and effective reparameterization of {IBM} model 2}.
\newblock In \emph{Proceedings of the 2013 Conference of the North {A}merican
  Chapter of the Association for Computational Linguistics: Human Language
  Technologies}, pages 644--648, Atlanta, Georgia. Association for
  Computational Linguistics.

\bibitem[{Filali and Bilmes(2005)}]{filali2005leveraging}
Karim Filali and Jeff Bilmes. 2005.
\newblock \href {https://doi.org/10.1109/ASRU.2005.1566493} {Leveraging
  multiple languages to improve statistical {MT} word alignments}.
\newblock In \emph{IEEE Workshop on Automatic Speech Recognition and
  Understanding, 2005.}, pages 92--97. IEEE.

\bibitem[{Garg et~al.(2019)Garg, Peitz, Nallasamy, and
  Paulik}]{garg-etal-2019-jointly}
Sarthak Garg, Stephan Peitz, Udhyakumar Nallasamy, and Matthias Paulik. 2019.
\newblock \href {https://doi.org/10.18653/v1/D19-1453} {Jointly learning to
  align and translate with transformer models}.
\newblock In \emph{Proceedings of the 2019 Conference on Empirical Methods in
  Natural Language Processing and the 9th International Joint Conference on
  Natural Language Processing (EMNLP-IJCNLP)}, pages 4453--4462, Hong Kong,
  China. Association for Computational Linguistics.

\bibitem[{ImaniGooghari et~al.(2021)ImaniGooghari, Jalili~Sabet, Dufter, Cysou,
  and Sch{\"u}tze}]{imanigooghari-etal-2021-parcoure}
Ayyoob ImaniGooghari, Masoud Jalili~Sabet, Philipp Dufter, Michael Cysou, and
  Hinrich Sch{\"u}tze. 2021.
\newblock \href {https://doi.org/10.18653/v1/2021.acl-demo.8} {{P}ar{C}our{E}:
  A parallel corpus explorer for a massively multilingual corpus}.
\newblock In \emph{Proceedings of the 59th Annual Meeting of the Association
  for Computational Linguistics and the 11th International Joint Conference on
  Natural Language Processing: System Demonstrations}, pages 63--72, Online.
  Association for Computational Linguistics.

\bibitem[{Jalili~Sabet et~al.(2020)Jalili~Sabet, Dufter, Yvon, and
  Sch{\"u}tze}]{jalili-sabet-etal-2020-simalign}
Masoud Jalili~Sabet, Philipp Dufter, Fran{\c{c}}ois Yvon, and Hinrich
  Sch{\"u}tze. 2020.
\newblock \href {https://doi.org/10.18653/v1/2020.findings-emnlp.147}
  {{S}im{A}lign: High quality word alignments without parallel training data
  using static and contextualized embeddings}.
\newblock In \emph{Findings of the Association for Computational Linguistics:
  EMNLP 2020}, pages 1627--1643, Online. Association for Computational
  Linguistics.

\bibitem[{Jeh and Widom(2002)}]{jeh2002simrank}
Glen Jeh and Jennifer Widom. 2002.
\newblock \href {https://doi.org/10.1145/775047.775126} {Simrank: a measure of
  structural-context similarity}.
\newblock In \emph{Proceedings of the eighth ACM SIGKDD international
  conference on Knowledge discovery and data mining}, pages 538--543.

\bibitem[{Joshi et~al.(2020)Joshi, Santy, Budhiraja, Bali, and
  Choudhury}]{joshi-etal-2020-state}
Pratik Joshi, Sebastin Santy, Amar Budhiraja, Kalika Bali, and Monojit
  Choudhury. 2020.
\newblock \href {https://doi.org/10.18653/v1/2020.acl-main.560} {The state and
  fate of linguistic diversity and inclusion in the {NLP} world}.
\newblock In \emph{Proceedings of the 58th Annual Meeting of the Association
  for Computational Linguistics}, pages 6282--6293, Online. Association for
  Computational Linguistics.

\bibitem[{Koehn et~al.(2003)Koehn, Och, and
  Marcu}]{koehn-etal-2003-statistical}
Philipp Koehn, Franz~J. Och, and Daniel Marcu. 2003.
\newblock \href {https://www.aclweb.org/anthology/N03-1017} {Statistical
  phrase-based translation}.
\newblock In \emph{Proceedings of the 2003 Human Language Technology Conference
  of the North {A}merican Chapter of the Association for Computational
  Linguistics}, pages 127--133.

\bibitem[{Koren et~al.(2009)Koren, Bell, and Volinsky}]{koren2009matrix}
Yehuda Koren, Robert Bell, and Chris Volinsky. 2009.
\newblock \href {https://doi.org/10.1109/MC.2009.263} {Matrix factorization
  techniques for recommender systems}.
\newblock \emph{Computer}, 42(8):30--37.

\bibitem[{Kuang et~al.(2015)Kuang, Choo, and Park}]{kuang2015nonnegative}
Da~Kuang, Jaegul Choo, and Haesun Park. 2015.
\newblock \href {https://doi.org/10.1007/978-3-319-09259-1_7} {Nonnegative
  matrix factorization for interactive topic modeling and document clustering}.
\newblock In \emph{Partitional Clustering Algorithms}, pages 215--243.
  Springer.

\bibitem[{Kumar et~al.(2007)Kumar, Och, and
  Macherey}]{kumar-etal-2007-improving}
Shankar Kumar, Franz~J. Och, and Wolfgang Macherey. 2007.
\newblock \href {https://www.aclweb.org/anthology/D07-1005} {Improving word
  alignment with bridge languages}.
\newblock In \emph{Proceedings of the 2007 Joint Conference on Empirical
  Methods in Natural Language Processing and Computational Natural Language
  Learning ({EMNLP}-{C}o{NLL})}, pages 42--50, Prague, Czech Republic.
  Association for Computational Linguistics.

\bibitem[{Lardilleux and Lepage(2008{\natexlab{a}})}]{lardilleux:hal-00368737}
Adrien Lardilleux and Yves Lepage. 2008{\natexlab{a}}.
\newblock \href {https://hal.archives-ouvertes.fr/hal-00368737} {{A truly
  multilingual, high coverage, accurate, yet simple, sub-sentential alignment
  method}}.
\newblock In \emph{{The 8th conference of the Association for Machine
  Translation in the Americas (AMTA 2008)}}, pages 125--132, Waikiki, Honolulu,
  United States.

\bibitem[{Lardilleux and
  Lepage(2008{\natexlab{b}})}]{lardilleux-lepage-2008-multilingual}
Adrien Lardilleux and Yves Lepage. 2008{\natexlab{b}}.
\newblock \href {https://www.aclweb.org/anthology/C08-2014} {Multilingual
  alignments by monolingual string differences}.
\newblock In \emph{Coling 2008: Companion volume: Posters}, pages 55--58,
  Manchester, UK. Coling 2008 Organizing Committee.

\bibitem[{Lardilleux and Lepage(2009)}]{lardilleux-lepage-2009-sampling}
Adrien Lardilleux and Yves Lepage. 2009.
\newblock \href {https://www.aclweb.org/anthology/R09-1040} {Sampling-based
  multilingual alignment}.
\newblock In \emph{Proceedings of the International Conference {RANLP}-2009},
  pages 214--218, Borovets, Bulgaria. Association for Computational
  Linguistics.

\bibitem[{Lewis and Xia(2008)}]{lewis-xia-2008-automatically}
William~D. Lewis and Fei Xia. 2008.
\newblock \href {https://www.aclweb.org/anthology/I08-2093} {Automatically
  identifying computationally relevant typological features}.
\newblock In \emph{Proceedings of the Third International Joint Conference on
  Natural Language Processing: Volume-{II}}.

\bibitem[{L{\"u} and Zhou(2010)}]{lu2010link}
Linyuan L{\"u} and Tao Zhou. 2010.
\newblock \href {https://doi.org/10.1209/0295-5075/89/18001} {Link prediction
  in weighted networks: The role of weak ties}.
\newblock \emph{EPL (Europhysics Letters)}, 89(1):18001.

\bibitem[{Luo et~al.(2014)Luo, Zhou, Xia, and Zhu}]{luo2014efficient}
Xin Luo, Mengchu Zhou, Yunni Xia, and Qingsheng Zhu. 2014.
\newblock \href {https://doi.org/10.1109/TII.2014.2308433} {An efficient
  non-negative matrix-factorization-based approach to collaborative filtering
  for recommender systems}.
\newblock \emph{IEEE Transactions on Industrial Informatics}, 10(2):1273--1284.

\bibitem[{Marchisio et~al.(2021)Marchisio, Xiong, and
  Koehn}]{marchisio2021embedding}
Kelly Marchisio, Conghao Xiong, and Philipp Koehn. 2021.
\newblock \href {https://arxiv.org/abs/2104.08721} {Embedding-enhanced
  {GIZA++}: Improving alignment in low-and high-resource scenarios using
  embedding space geometry}.
\newblock \emph{arXiv preprint arXiv:2104.08721}.

\bibitem[{Mayer and Cysouw(2012)}]{mayer-cysouw-2012-language}
Thomas Mayer and Michael Cysouw. 2012.
\newblock \href {https://www.aclweb.org/anthology/W12-0209} {Language
  comparison through sparse multilingual word alignment}.
\newblock In \emph{Proceedings of the {EACL} 2012 Joint Workshop of {LINGVIS}
  {\&} {UNCLH}}, pages 54--62, Avignon, France. Association for Computational
  Linguistics.

\bibitem[{Mayer and Cysouw(2014)}]{mayer-cysouw-2014-creating}
Thomas Mayer and Michael Cysouw. 2014.
\newblock \href
  {http://www.lrec-conf.org/proceedings/lrec2014/pdf/220_Paper.pdf} {Creating a
  massively parallel {B}ible corpus}.
\newblock In \emph{Proceedings of the Ninth International Conference on
  Language Resources and Evaluation ({LREC}'14)}, pages 3158--3163, Reykjavik,
  Iceland. European Language Resources Association (ELRA).

\bibitem[{Melamed(1998)}]{melamed1998blinker}
I.~Dan Melamed. 1998.
\newblock \href {http://arxiv.org/abs/cmp-lg/9805005} {Manual annotation of
  translational equivalence: The blinker project}.
\newblock \emph{CoRR}, cmp-lg/9805005.

\bibitem[{M{\"u}ller(2017)}]{muller-2017-treatment}
Mathias M{\"u}ller. 2017.
\newblock \href {https://doi.org/10.18653/v1/W17-4804} {Treatment of markup in
  statistical machine translation}.
\newblock In \emph{Proceedings of the Third Workshop on Discourse in Machine
  Translation}, pages 36--46, Copenhagen, Denmark. Association for
  Computational Linguistics.

\bibitem[{Nakov and Ng(2012)}]{nakov2012improving}
Preslav Nakov and Hwee~Tou Ng. 2012.
\newblock \href {https://doi.org/10.1613/jair.3540} {Improving statistical
  machine translation for a resource-poor language using related resource-rich
  languages}.
\newblock \emph{Journal of Artificial Intelligence Research}, 44:179--222.

\bibitem[{Och and Ney(2003{\natexlab{a}})}]{och03:asc}
Franz~Josef Och and Hermann Ney. 2003{\natexlab{a}}.
\newblock \href {https://doi.org/10.1162/089120103321337421} {A systematic
  comparison of various statistical alignment models}.
\newblock \emph{Computational Linguistics}, 29(1).

\bibitem[{Och and Ney(2003{\natexlab{b}})}]{och-ney-2003-systematic}
Franz~Josef Och and Hermann Ney. 2003{\natexlab{b}}.
\newblock \href {https://doi.org/10.1162/089120103321337421} {A systematic
  comparison of various statistical alignment models}.
\newblock \emph{Computational Linguistics}, 29(1):19--51.

\bibitem[{{\"{O}}stling(2014)}]{ostling2014bayesian}
Robert {\"{O}}stling. 2014.
\newblock \href {https://doi.org/10.3115/v1/e14-4024} {Bayesian word alignment
  for massively parallel texts}.
\newblock In \emph{Proceedings of the 14th Conference of the European Chapter
  of the Association for Computational Linguistics, {EACL} 2014, April 26-30,
  2014, Gothenburg, Sweden}, pages 123--127. The Association for Computer
  Linguistics.

\bibitem[{{\"O}stling(2015)}]{ostling-2015-word}
Robert {\"O}stling. 2015.
\newblock \href {https://doi.org/10.3115/v1/P15-2034} {Word order typology
  through multilingual word alignment}.
\newblock In \emph{Proceedings of the 53rd Annual Meeting of the Association
  for Computational Linguistics and the 7th International Joint Conference on
  Natural Language Processing (Volume 2: Short Papers)}, pages 205--211,
  Beijing, China. Association for Computational Linguistics.

\bibitem[{{\"O}stling and Tiedemann(2016)}]{ostling2016efficient}
Robert {\"O}stling and J{\"o}rg Tiedemann. 2016.
\newblock \href
  {https://content.sciendo.com/downloadpdf/journals/pralin/106/1/article-p125.xml}
  {Efficient word alignment with {M}arkov {C}hain {M}onte {C}arlo}.
\newblock \emph{The Prague Bulletin of Mathematical Linguistics}, 106(1).

\bibitem[{Rasooli et~al.(2018)Rasooli, Farra, Radeva, Yu, and
  McKeown}]{rasooli2018cross}
Mohammad~Sadegh Rasooli, Noura Farra, Axinia Radeva, Tao Yu, and Kathleen
  McKeown. 2018.
\newblock \href {https://doi.org/10.1007/s10590-017-9202-6} {Cross-lingual
  sentiment transfer with limited resources}.
\newblock \emph{Machine Translation}, 32(1):143--165.

\bibitem[{Rothe and Sch{\"u}tze(2014)}]{rothe-schutze-2014-cosimrank}
Sascha Rothe and Hinrich Sch{\"u}tze. 2014.
\newblock \href {https://doi.org/10.3115/v1/P14-1131} {{C}o{S}im{R}ank: A
  flexible {\&} efficient graph-theoretic similarity measure}.
\newblock In \emph{Proceedings of the 52nd Annual Meeting of the Association
  for Computational Linguistics (Volume 1: Long Papers)}, pages 1392--1402,
  Baltimore, Maryland. Association for Computational Linguistics.

\bibitem[{Sch{\"u}tze(1998)}]{schutze-1998-automatic}
Hinrich Sch{\"u}tze. 1998.
\newblock \href {https://www.aclweb.org/anthology/J98-1004} {Automatic word
  sense discrimination}.
\newblock \emph{Computational Linguistics}, 24(1):97--123.

\bibitem[{Shahnaz et~al.(2006)Shahnaz, Berry, Pauca, and
  Plemmons}]{shahnaz2006document}
Farial Shahnaz, Michael~W Berry, V~Paul Pauca, and Robert~J Plemmons. 2006.
\newblock \href {https://doi.org/10.1016/j.ipm.2004.11.005} {Document
  clustering using nonnegative matrix factorization}.
\newblock \emph{Information Processing \& Management}, 42(2):373--386.

\bibitem[{Shi et~al.(2021)Shi, Zettlemoyer, and Wang}]{shi-etal-2021-bilingual}
Haoyue Shi, Luke Zettlemoyer, and Sida~I. Wang. 2021.
\newblock \href {https://doi.org/10.18653/v1/2021.acl-long.67} {Bilingual
  lexicon induction via unsupervised bitext construction and word alignment}.
\newblock In \emph{Proceedings of the 59th Annual Meeting of the Association
  for Computational Linguistics and the 11th International Joint Conference on
  Natural Language Processing (Volume 1: Long Papers)}, pages 813--826, Online.
  Association for Computational Linguistics.

\bibitem[{Tavakoli and Faili(2014)}]{tavakoli2014phrase}
Leila Tavakoli and Heshaam Faili. 2014.
\newblock \href {https://www.sid.ir/en/VEWSSID/J_pdf/125620140307.pdf} {Phrase
  alignments in parallel corpus using bootstrapping approach}.
\newblock \emph{International Journal of Information \& Communication
  Technology Research}, 6(3).

\bibitem[{Wu and Dredze(2020)}]{wu-dredze-2020-explicit}
Shijie Wu and Mark Dredze. 2020.
\newblock \href {https://doi.org/10.18653/v1/2020.emnlp-main.362} {Do explicit
  alignments robustly improve multilingual encoders?}
\newblock In \emph{Proceedings of the 2020 Conference on Empirical Methods in
  Natural Language Processing (EMNLP)}, pages 4471--4482, Online. Association
  for Computational Linguistics.

\bibitem[{Xu et~al.(2003)Xu, Liu, and Gong}]{xu2003document}
Wei Xu, Xin Liu, and Yihong Gong. 2003.
\newblock \href {https://doi.org/10.1145/860435.860485} {Document clustering
  based on non-negative matrix factorization}.
\newblock In \emph{Proceedings of the 26th annual international ACM SIGIR
  conference on Research and development in informaion retrieval}, pages
  267--273.

\bibitem[{Yli-Jyr{\"a} et~al.(2020)Yli-Jyr{\"a}, Purhonen, Liljeqvist, Antturi,
  Nieminen, R{\"a}ntil{\"a}, and Luoto}]{yli-jyra-etal-2020-helfi}
Anssi Yli-Jyr{\"a}, Josi Purhonen, Matti Liljeqvist, Arto Antturi, Pekka
  Nieminen, Kari~M. R{\"a}ntil{\"a}, and Valtter Luoto. 2020.
\newblock \href {https://www.aclweb.org/anthology/2020.lrec-1.522} {{HELFI}: a
  {H}ebrew-{G}reek-{F}innish parallel {B}ible corpus with cross-lingual
  morpheme alignment}.
\newblock In \emph{Proceedings of the 12th Language Resources and Evaluation
  Conference}, pages 4229--4236, Marseille, France. European Language Resources
  Association.

\bibitem[{Zamani et~al.(2016)Zamani, Dadashkarimi, Shakery, and
  Croft}]{zamani2016pseudo}
Hamed Zamani, Javid Dadashkarimi, Azadeh Shakery, and W~Bruce Croft. 2016.
\newblock \href {https://doi.org/10.1145/2983323.2983844} {Pseudo-relevance
  feedback based on matrix factorization}.
\newblock In \emph{Proceedings of the 25th ACM international on conference on
  information and knowledge management}, pages 1483--1492.

\bibitem[{Zenkel et~al.(2020)Zenkel, Wuebker, and
  DeNero}]{zenkel-etal-2020-end}
Thomas Zenkel, Joern Wuebker, and John DeNero. 2020.
\newblock \href {https://doi.org/10.18653/v1/2020.acl-main.146} {End-to-end
  neural word alignment outperforms {GIZA}++}.
\newblock In \emph{Proceedings of the 58th Annual Meeting of the Association
  for Computational Linguistics}, pages 1605--1617, Online. Association for
  Computational Linguistics.

\bibitem[{Zhang and Chen(2018)}]{zhang2018link}
Muhan Zhang and Yixin Chen. 2018.
\newblock \href {https://dl.acm.org/doi/pdf/10.5555/3327345.3327423} {Link
  prediction based on graph neural networks}.
\newblock \emph{Advances in Neural Information Processing Systems},
  31:5165--5175.

\bibitem[{Zhao and Gildea(2010)}]{zhao-gildea-2010-fast}
Shaojun Zhao and Daniel Gildea. 2010.
\newblock \href {https://www.aclweb.org/anthology/D10-1058} {A fast fertility
  hidden {M}arkov model for word alignment using {MCMC}}.
\newblock In \emph{Proceedings of the 2010 Conference on Empirical Methods in
  Natural Language Processing}, pages 596--605, Cambridge, MA. Association for
  Computational Linguistics.

\bibitem[{Zhou et~al.(2009)Zhou, L{\"u}, and Zhang}]{zhou2009predicting}
Tao Zhou, Linyuan L{\"u}, and Yi-Cheng Zhang. 2009.
\newblock \href {https://doi.org/10.1140/epjb/e2009-00335-8} {Predicting
  missing links via local information}.
\newblock \emph{The European Physical Journal B}, 71(4):623--630.

\end{thebibliography}

\appendix

\section{Pipeline Details}
\label{sec:appendix_pipe}
\begin{figure}[h]
	\centering
	\includegraphics[width=1.0\linewidth]{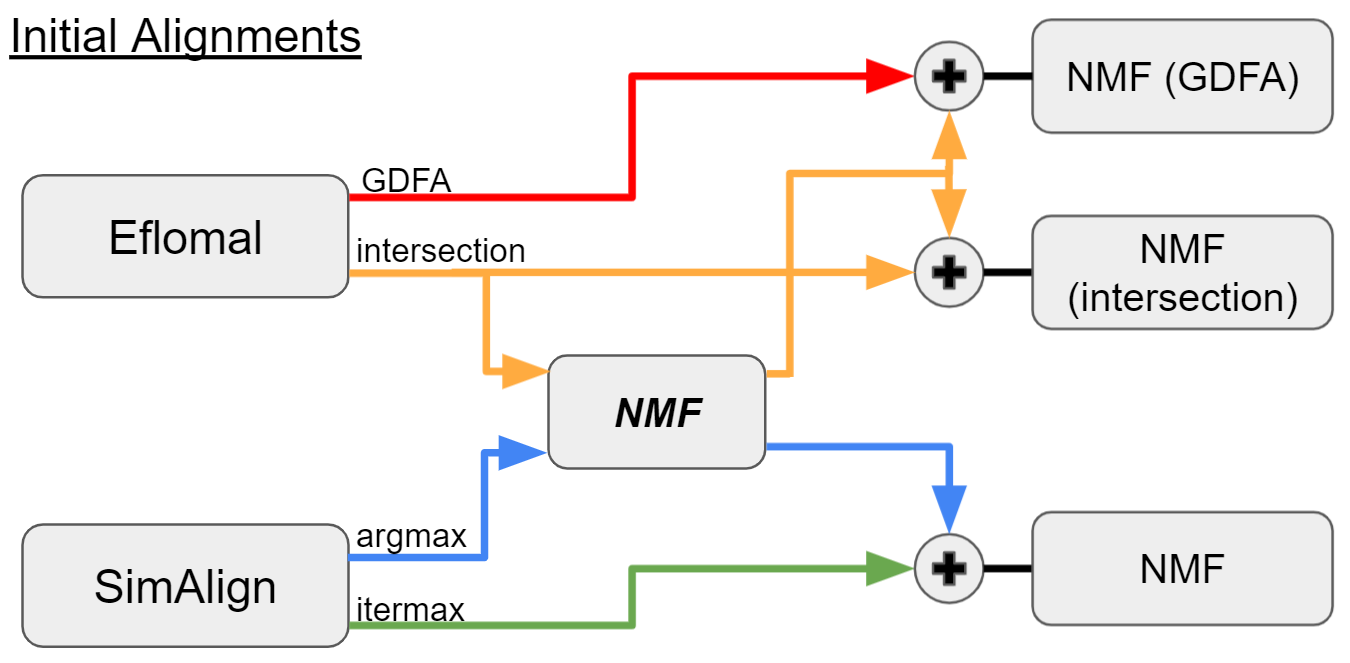}
	\caption{The pipeline for NMF alignments}
\label{fig:pipeline_nmf}
\end{figure}

\begin{figure}[h]
	\centering
	\includegraphics[width=1.0\linewidth]{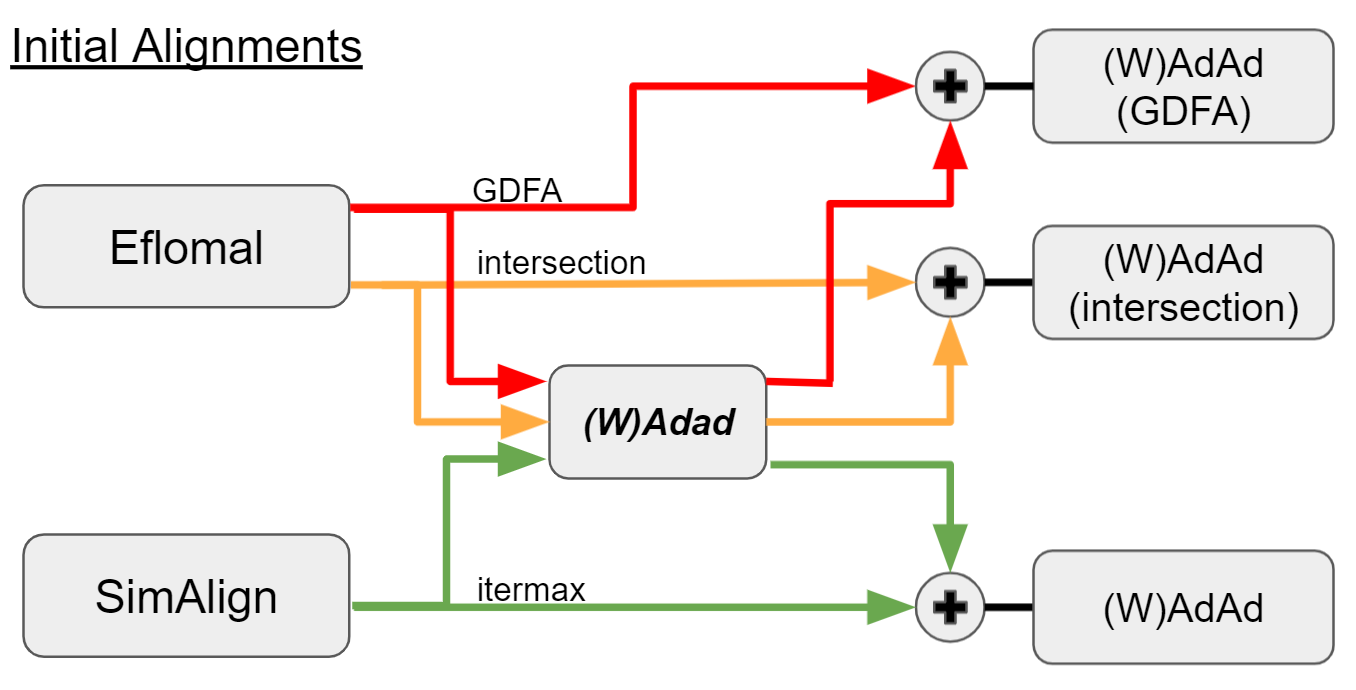}
	\caption{The pipeline for AdAd and WAdAd alignments}
        \label{fig:pipeline_adad}
\end{figure}

There are several elements of the
\framework{} pipeline that
can be configured by the user, e.g., depending on
whether precision or recall are more important for an application.
Here we show 
in Figures~\ref{fig:pipeline_nmf} and
\ref{fig:pipeline_adad} the two pipeline configurations we
used for the results in the paper.


\end{document}